\documentclass[conference, a4paper]{IEEEtran}
\usepackage{cite}
\usepackage[pdftex]{graphicx}
\usepackage[cmex10]{amsmath}
\usepackage{algorithm}
\usepackage{algorithmicx}
\usepackage{algpseudocode}
\usepackage{mdwmath}
\usepackage{mdwtab}
\usepackage{amssymb}
\usepackage{amsmath}
\usepackage{subfigure}
\usepackage{tabularx}
\usepackage{mdwlist}
\usepackage{url}
\usepackage{multirow}
\usepackage{latexsym}
\usepackage{graphicx}
\usepackage{booktabs}
\usepackage{csquotes}
\usepackage{graphicx}
\usepackage[table,xcdraw]{xcolor}

\ifCLASSINFOpdf
\else
\fi

\newcommand{\comment}[1]{}

\DeclareMathAlphabet{\mathitbf}{OML}{cmm}{b}{it}

\newcommand{\ie}{\emph{i.e.}}
\newcommand{\etal}{\emph{et~al.}}
\renewcommand{\Huge}{\fontsize{23.5pt}{27pt}\selectfont}
\newcommand{\sq}{\hbox{\rlap{$\sqcap$}$\sqcup$}}
\newcommand{\qed}{\hspace*{\fill}\sq}

\begin{document}
\title{KoreALBERT: Pretraining a Lite BERT Model for Korean Language Understanding}

\author{
\IEEEauthorblockN{Hyunjae Lee, Jaewoong Yoon, Bonggyu Hwang, Seongho Joe, Seungjai Min, Youngjune Gwon}
\IEEEauthorblockA{Samsung SDS}
}

\maketitle

\begin{abstract}
A Lite BERT (ALBERT) has been introduced to scale up deep bidirectional representation learning for natural languages. Due to the lack of pretrained ALBERT models for Korean language, the best available practice is the multilingual model or resorting back to the any other BERT-based model. In this paper, we develop and pretrain KoreALBERT, a monolingual ALBERT model specifically for Korean language understanding. We introduce a new training objective, namely Word Order Prediction (WOP), and use alongside the existing MLM and SOP criteria to the same architecture and model parameters. Despite having significantly fewer model parameters (thus, quicker to train), our pretrained KoreALBERT outperforms its BERT counterpart on 6 different NLU tasks. Consistent with the empirical results in English by Lan et al., KoreALBERT seems to improve downstream task performance involving multi-sentence encoding for Korean language. The pretrained KoreALBERT is publicly available to encourage research and application development for Korean NLP.
\end{abstract}

\IEEEpeerreviewmaketitle

\section{Introduction}
Pre-trained language models are becoming an essential component to build a modern natural language processing (NLP) application. Previously, recurrent neural nets such as LSTM have dominated sequence-to-sequence (seq2seq)~\cite{seq2seq} modeling for natural languages, upholding state-of-the-art performances for core language understanding tasks. Since the introduction of the Transformer~\cite{NIPS2017_7181}, recurrent structures in a neural language model are reconsidered and opted for attention, a mechanism that relates different positions in a sequence to compute a representation of the sequence.

Devlin~\etal~\cite{bert} have proposed Bidirectional Encoder Representations from Transformers (BERT) to improve on predominantly unidirectional training of a language model by using the masked language model (MLM) training objective. MLM is an old concept dating back to the 1950s~\cite{taylor}. By jointly conditioning on both left and right context in all layers, the MLM objective has made pre-training of the deep bidirectional language encoding possible. BERT uses an additional loss for pre-training known as next-sentence prediction (NSP). NSP is designed to learn high-level linguistic coherence by predicting whether or not given two text segments should appear consecutively as in the original text. NSP can improve performance on downstream NLP tasks such as natural language inference that would require reasoning about inter-sentence relations.

A Lite BERT (ALBERT) uses parameter reduction techniques to alleviate scaling problems for BERT. ALBERT's cross-layer parameter sharing can be thought as a form of regularization that helps stabilize the pre-training and generalize despite the substantially reduced number of model parameters. Also, the sentence order prediction (SOP) objective in ALBERT replaces the ineffective the next sentence prediction (NSP) loss in BERT for better inter-sentence coherence. 

Downstream tasks play critical measures for evaluating emerging language models and NLP applications today. Pre-trained language models are central to downstream task evaluations such as machine translation, text classification, and machine reading comprehension. At a high level, there are two approaches to use pre-trained language models. First, pre-trained models can provide additional feature representations for a downstream task. More importantly, pre-trained models can be a baseline upon which the downstream task is fine-tuned. 

By having an expensive, but shareable pre-training followed by much smaller fine-tuning, it is a powerful paradigm to focus on optimizing the performance of a downstream NLP task. Self-supervised learning with large corpora allows a suitable starting point for an outer task-specific layer being optimized from scratch while reusing the pre-trained model parameters. 

Since its introduction, BERT has achieved state-of-the-art accuracy performances for natural language understanding tasks such as GLUE~\cite{GLUE}, MultiNLI~\cite{MultiNLI}, SQuAD v1.1~\cite{squad} \& SQuAD v2.0~\cite{squad2}, and CoNLL-2003 NER~\cite{conll}. Despite having fewer parameters than BERT, ALBERT has been able to achieve new state-of-the-art results on the GLUE, RACE~\cite{race}, and SQuAD benchmarks. 

It is important to remark that a large network is crucial in pushing state-of-the-art results for downstream tasks. While BERT gives a sound choice to build a general language model trained on large corpora, it is difficult to experiment with training large BERT models due to the memory limitations and computational constraints. Training BERT-large in fact is a lengthy process of consuming significant hardware resources. Besides, there are already a wide variety of languages pre-trained in BERT, which include the multilingual BERT and monolingual models pre-trained in 104 different languages. ALBERT, however, gives a much narrower choice in languages. 

Asserting an argument that having a better language model is roughly equivalent to pre-train a large model, all without imposing too much memory and computational requirements, we choose to go with ALBERT. In this paper, we develop and train KoreALBERT, a monolingual ALBERT model for Korean language understanding. Compared to a multilingual model, monolingual language models are known to optimize the performance for a specific language in every aspect, including downstream tasks critical to build modern NLP systems and applications. 

In addition to the original ALBERT MLM and SOP training objectives, we introduce a word order prediction (WOP) loss. WOP is fully compatible with the MLM and SOP losses and can be added gracefully in implementation. Our pre-trained KoreALBERT could outperform multilingual BERT and its BERT counterpart on a brief evaluation with KorQuAD 1.0 benchmark for machine reading comprehension. Consistent with the empirical results of ALBERT pre-trained in English reported by Lan~\etal~\cite{albert}, KoreALBERT seems to improve supervised downstream task performances involving multiple Korean sentences. 

The rest of this paper is organized as follows. In Section II, we provide background on pre-trained neural language models. Section III presents KoreALBERT. In Section IV, we describe our implementation, pre-training, and empirical evaluation of KoreALBERT. Section V concludes the paper. Our pre-trained KoreALBERT is publicly available to encourage NLP research and application development for Korean language. 

\section{Background}
\subsection{Transformer, BERT, and ALBERT}
Transformer~\cite{NIPS2017_7181} is a sequence transduction model based solely on attention mechanism, skipping any recurrent and convolutional structures of a neural network. The transformer architecture includes multiple identical encoder and decoder blocks stacked on top of each other. While the encoder captures linguistic information of the input sequence and produces the contextual representations, the decoder generates output sequence corresponding to its pair of input. Thanks to multi-head self-attention layers in an encoder block, transformer can acquire varying attentions within a single sequence and alleviate inevitable dragging caused during the training of a recurrent neural network.

BERT distinguishes itself from other language models that predict the next word given previous words by introducing new training methods. Instead of predicting the next token given only previous tokens, it has to predict replaced word by special token \texttt{[MASK]}. This training strategy gives BERT bidirectionality which means having an access to left and right context around the target word. Thus, BERT can produce deep bidirectional representation of input sequence.

RoBERTa~\cite{ROBERTA}, ALBERT~\cite{albert} and other variants~\cite{SpanBERT, BERT-WWM} utilize bidirectional context representation and established state-of-the-art results on a wide range of NLP tasks. BERT is trained with the masked language modeling (MLM) and the next sentence prediction (NSP) losses. NSP is a binary classification task to predict whether or not given two segments separated by another special token \texttt{[SEP]} follow each other in the original text. The task is intended to learn the relationship between two sentences in order to use on many downstream tasks of which input template consists of two sentences as in question answering (QA) and sentence entailment~\cite{bert}.

Recently, there is a criticism toward NSP that the NSP loss does not necessarily help improve the downstream task performances~\cite{albert, ROBERTA, XLNET} for its loose inter-sentential coherence. Among them, ALBERT, whose architecture is derived from BERT, uses a sentence order prediction(SOP) task instead. In the SOP task, negative examples consist of a pair of sentences from the same document, but the sentence order is swapped, and the model should predict whether or not the order is swapped.
With the improved SOP loss and other parameter reduction techniques, ALBERT significantly reduces the number of parameters--\ie,~18x fewer for BERT-large, while achieving similar or better performance on downstream tasks~\cite{albert}. KoreALBERT takes the unmodified ALBERT architecture as a baseline. We train KoreALBERT from scratch on large Korean corpora collected online. 

\subsection{Related Work}
Google has released BERT multilingual model (M-BERT) pre-trained using 104 different languages including the Korean. Karthikeyan~\etal~\cite{k2019crosslingual} show why and how well M-BERT works on many downstream NLP tasks without explicitly training with monolingual corpus. 
More recently, Facebook AI Research presented crosslingual model (XLM-R)~\cite{xlm-r} generally outperforming M-BERT. 
Recent literature argues that a monolingual language model is consistently superior to M-BERT. For French, FlauBERT~\cite{flaubert} and CamemBERT~\cite{camembert} with the same approach as RoBERTa have been released. ALBERTo~\cite{alberto} focuses on Italian social network data. BERTje~\cite{bertje} for Dutch and FinBERT~\cite{FinBERT} for Finnish have been developed. They both have achieved superior results on the majority of downstream NLP tasks compared to M-BERT.

Some previous work in the Korean language has focused on learning static representations by using language-specific properties~\cite{subword}.
More recently, SKT Brain has released BERT \footnote{https://github.com/SKTBrain/KoBERT} and GPT-2 pre-trained on large Korean corpora.\footnote{https://github.com/SKT-AI/KoGPT2}
Korean Electronics and Telecommunications Research Institute (ETRI) has released two versions of BERT: the morpheme analytic based and the syllable based model.\footnote{http://aiopen.etri.re.kr/service\_dataset.php} These models are worthwhile to experiment with and provide good benchmark evaluations in Korean language model research.

BART~\cite{BART} features interesting denoising approaches for input text used in pre-training such as sentence permutation and text infilling. In the sentence permutation task, an input document is divided into sentences and shuffled in a random order. A combination of text infilling and sentence shuffling tasks has shown significant improvement of the performance over either applied separately. Inspired by BART, we have formulated word order prediction (WOP), a new pre-training loss used alongside the MLM and SOP losses for KoreALBERT. Differentiated from BART, which is essentially a sentence-level shuffling, WOP is an intra-sentence, token-level shuffling.
\section{KoreALBERT: Training Korean Language Model Using ALBERT}
\subsection{Architecture}
KoreALBERT is a multi-layer bidirectional Transformer encoder with the same factorized embedding parameterization and cross-layer sharing as ALBERT. Inheriting ALBERT-base, KoreALBERT-base has 12 parameter sharing layers with an embedding size of 128 dimensions, 768 hidden units, 12 heads, and GELU nonlinearities~\cite{GELUs}. The total number of parameters in KoreALBERT-base is 12 millions, and it increases to 18-million parameters for KoreALBERT-large having 1024 hidden dimensions. 

Lan~\etal~\cite{albert} argues that removing dropout has significantly helped pretraining with the masked language modeling (MLM) loss. For KoreALBERT, however, we have made an empirical decision to keep dropout after observing degraded downstream performances without dropout. 

\subsection{Training Objectives} 
ALBERT pretrains on two objectives: masked language modeling (MLM) and sentence order prediction (SOP) losses. We keep both objectives for KoreALBERT and introduce an additional training objective called word order prediction (WOP). 

\textbf{Word Order Prediction (WOP).} Korean is an agglutinative language that a combination of affixes and word roots determines usage and meaning~\cite{song}. Decomposing a Korean word into several morphemes and shuffling its order can introduce grammatical errors and semantic altercations. We impose a word order prediction (WOP) loss for pretraining KoreALBERT. The WOP objective is a cross-entropy loss on predicting a correct order of shuffled tokens. 

WOP is fully compatible with the ALBERT MLM and SOP, and we expect to reinforce correct agglutination (or point out incorrect agglutinative usages) beyond simply checking intra-sentence word orderings. There is an interesting point of view about WOP mixed with MLM and SOP towards the problem of generating a full sentence from a small subset of permuted words. Our primary focus of this paper is on the empirical side of the design and pretraining of an ALBERT-based foreign language model rather than a formal analysis on training objectives.  

\begin{figure*}[!t]
\centering
\includegraphics[width=.8\textwidth]{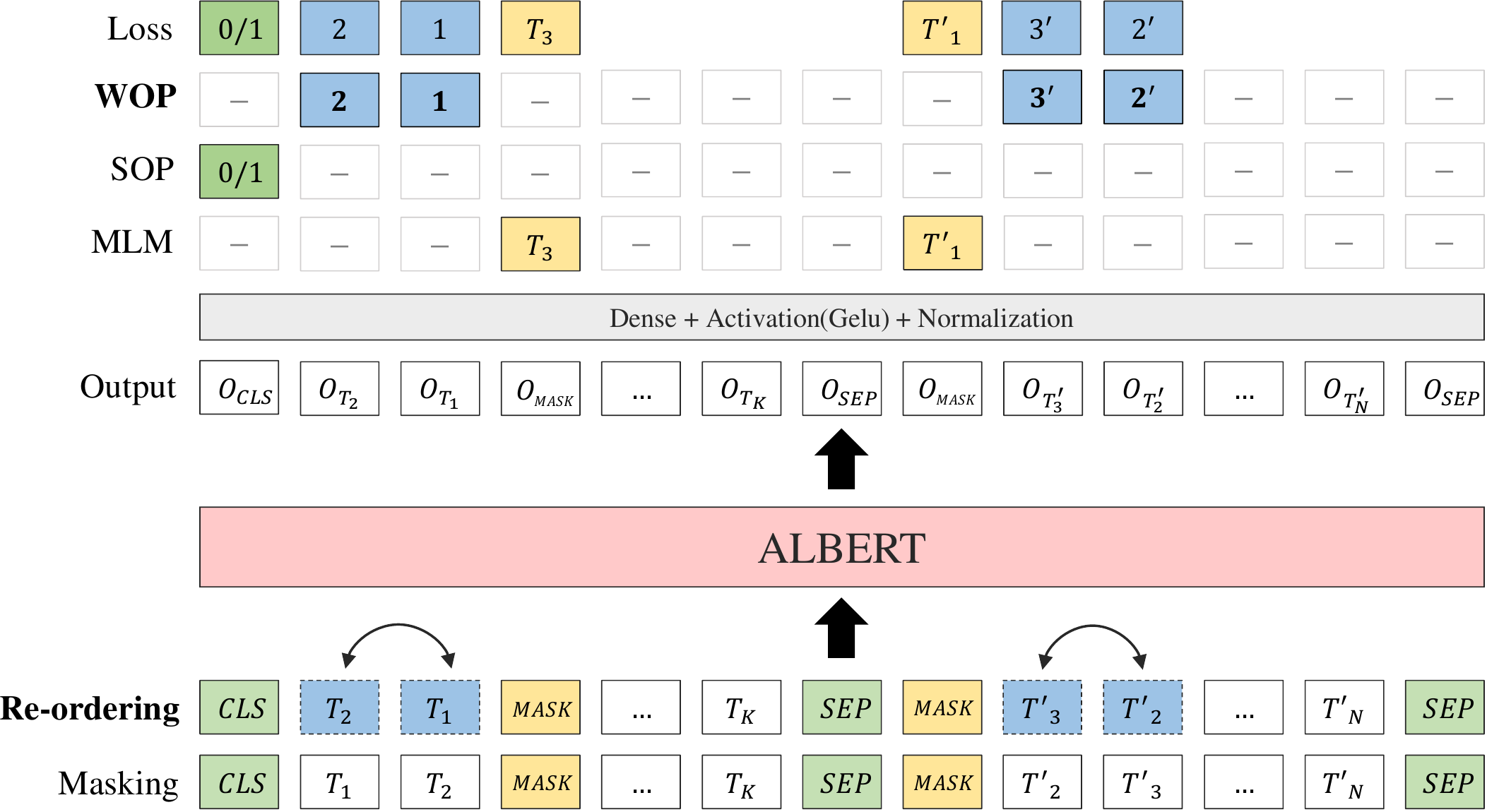}
\caption{Pre-training KoreALBERT with the MLM, SOP, and WOP objectives. The loss (on top) with respect to all three objectives is calculated for illustrative purposes. In our implementation, classification layer (highlighted gray) in the middle consisting of three identical heads produces a logit vector with respect to each label.}
\label{fig:wop}
\end{figure*}

The pretraining of KoreALBERT is illustrated in Fig.~\ref{fig:wop}. A randomly sampled subset of tokens in the input text are replaced with \texttt{[MASK]}. MLM computes a cross-entropy loss on prediction of the masked tokens. As with ALBERT-base, we uniformly choose 15\% of the input tokens for possible masking, and the 80\% of the chosen are actually replaced with \texttt{[MASK]}, leaving 10\% unchanged and the rest replaced with randomly selected tokens. SOP is known to focus on modeling inter-sentence coherence. The SOP loss uses two consecutive segments from the same text as a positive example and as a negative example if their order is swapped. We have found that if WOP is too difficult, it can crucially impact the KoreALBERT performance on downstream evaluations. 

We have experimentally determined WOP to inter-work with MLM and SOP and limited the shuffling rate up to 15\%, which seemingly realizes the best empirical performance for our case. In addition, we have decided to include WOP into only specific portion of all batches. We revisit more detailed description of our experimental setup in Section 4. Like MLM, we choose a uniformly random set of tokens for WOP. The most crucial part of integrating WOP into pretraining is \emph{not} switching tokens across \texttt{[MASK]}. This constraint minimizes the corruption of contextual bidirectionality that acts as essential information in denoising the \texttt{[MASK]} tokens.

\subsection{Optimization}
We use the LAMB optimizer~\cite{76Minutes} with a learning rate of $1.25\times 10^{-3}$ and a warm-up ratio $1.25\times 10^{-2}$. To speed up the pretraining, we maintain an input sequence length of 128 tokens despite the risk of suboptimal performance. Due to memory limitations, it is necessary to use gradient accumulation steps for a batch size of 2,048, which is comparable to BERT. We apply a dropout rate of 0.1 on all layers and attention weights. We use a GELU activation function~\cite{GELUs}.

\section{Experiments}
\subsection{Implementation}
We implement KoreALBERT based on Hugging Face's transformer library~\cite{huggingface} with almost an identical model configuration for ALBERT-base. We add another linear classifier on top of the encoder output for WOP task. The added layer is used to predict the probability of the original position of words in the sequence via softmax. Like the MLM objective, we take into account only switched tokens to compute the cross-entropy loss. We train our model using 4 NVidia V100 GPUs with half-precision floating-point weights.

\subsection{Data}
Many BERT-style language models include Wikipedia in the pre-training corpora for a wide coverage of topics in relatively high-quality writing. Korean Wikipedia currently ranks the 23rd by volume, and this is just 7.8\% compared to English Wikipedia. To supplement training examples and the diversity of our corpus, we also use the text from NamuWiki\footnote{https://en.wikipedia.org/wiki/Namuwiki}, which is another Korean online encyclopedia that contains more subjective opinions covering a variety of topics and writing styles.

\subsubsection{Pretraining corpora}
our pretraining corpora include the following.
\begin{itemize}
\item Web News: all articles from 8 major newspapers of Korea accross the topics including politics, social, economics, culture, IT, opinion, and sports from January 1, 2007 to December 31, 2019.
\item Korean Wikipedia: 490,220 documents crawled in October, 2019.
\item NamuWiki: 740,094 documents crawled in December, 2019.
\item Book corpus: plots and editorial reviews about all Korean books published in 2010 to December 31, 2019\footnote{http://book.interpark.com/} 
\end{itemize}

\subsubsection{Text preprocessing} 
We have preprocessed our text data in the following manner. First, we remove all meta-tags such as the date of writing and name(s) of the author(s) in newspapers appearing in the beginning and at the end of each article. We think that the meta-tags do not contain any contextual or semantic information essential for NLU tasks. We also adjust the proportion of categories making up the news corpus in order to avoid topical bias of the examples. We tokenize the corpora into subwords using SentencePiece tokenizer~\cite{sentencepiece} like ALBERT to construct vocabulary of a size 32k. We mask randomly sampled 15\% of the words using the whole word masking strategy recently introduced by BERT. After cleaning and regularizing text, we obtain 43GB text with 325 million sentences, which are equivalent to 4.4 billion words or 18 billion characters.

\subsection{Compatibility of Word Order Prediction (WOP)}
We have performed ablation experiments with and without WOP to empirically observe its compatibility with the MLM and SOP objectives by pretraining for 125K steps, which is the half of the entire pre-training. A critical decision to introduce new noise via WOP is how many training examples should entail the additional noising process as well as how many tokens should be shuffled inside a sentence. We sample batches to contain re-ordered tokens proportionally from 30 to 100\%. We have observed that about 30\--50\% shuffling achieves a good performance for most cases. Results are averaged over 10 different seeds and summarized in Table~\ref{tab:portion}.

We set up three combinations of the pretraining objectives to compare against one another in the downstream evaluations to highlight the effect of WOP. We also observe the intrinsic performance of each objective. In the WOP and MLM combination, we configure the portion of corrupted examples to 30\% for the WOP objective. From the result averaged over 10 different seeds in Table~\ref{tab:performance_wop}, WOP hardly hurts the performance of MLM or SOP. The accuracy of MLM and WOP tasks has improved in case of leaving the SOP objective out. We believe that the best usage for WOP is not to disturb other intrinsic tasks for pretraining. WOP should be added by carefully observing the performance of other objectives on different WOP configurations.

As expected, the deletion of SOP has caused a degradation more than 3\% in the downstream performances of semantic textual similarity (8,628 examples) and paraphrase detection (7,576 examples). These two tasks are relatively small data experiments. Surprisingly, the performance of KorNLI is better without SOP because NLI tasks depend on inter-sentence coherence. Note that KorNLI is a much larger dataset (950,354 examples) compared to the semantic textual similarity and paraphrase detection datasets. Combining the two denoising objectives MLM and WOP seems to alleviate the performance degradation for a classification task with multi-sentence input.

\begin{table*}[t]
\centering
\caption{Experimental results on downstream tasks according to different portion of word order prediction tasks}
\label{tab:portion}
\begin{tabularx}{0.58\textwidth}{c|cccccc}
\toprule
\multirow{2}{*}{Portion of WOP} &  KorNLI &  KorSTS&  NSMC&  PD &  NER & KorQuAD1.0 \\
									   &  acc &  spearman &  acc &  acc &  acc &  f1  \\
\midrule 
100 \% & \bf 76.8 & 74.8 & 88.3 & 92.3 & 80.6 & 89.4 \\
50 \%& 76.4 & \bf 76.6 & 88.3 & 92.7 & \bf 81.2 & 89.3 \\
30 \%& 76.6 & 75.4 & \bf 88.4 & \bf 93.2 & 80.7 & \bf 89.8 \\
\bottomrule
\end{tabularx}
\end{table*}

\begin{table*}[t]
\centering
\caption{Experimental results on Downstream task performance comparing between different combination of pretraining objectives}
\label{tab:performance_wop}
\begin{tabularx}{0.78\textwidth}{l|ccc|cccccc}
\toprule
\multirow{2}{*} { Objectives} &  MLM & SOP&  WOP &  KorNLI &  KorSTS&  NSMC&  PD & NER & KorQuAD1.0  \\
								  &  acc &  acc &  acc &  acc &  spearman &  acc &  acc &  acc &  f1  \\
\midrule
MLM + SOP & 35.3 & 79.8 & - & 76.4 & 75.6 & \bf 88.6 & 92.9 & 80.7 & 89.5 \\
MLM + SOP + WOP & 35.1 & 79.1 & 80.7 & \bf 76.9 & \bf 76.6 & 88.4 & \bf 93.2 & \bf 81.2 & \bf 89.8\\
MLM + WOP & 35.6 &- & 84.0 & 76.8 & 73.3 & 88.5 & 92.3 & 81.0 & 89.3 \\
\bottomrule
\end{tabularx}
\end{table*}

\begin{table*}[t]
\centering
\caption{Experimental results on downstream tasks and model parameters}
\label{tab:final_perform}
\begin{tabularx}{0.8\textwidth}{l|cc|cccccc|c}
\toprule
\multirow{2}{*} {Model} & \multirow{2}{*}{Params} & \multirow{2}{*}{Speedup} & KorNLI &  KorSTS&  NSMC&  PD &  NER &  KorQuAD1.0 & \multirow{2}{*}{ Avg.} \\
							 &&& acc           & spearman &         acc  &      acc &       acc &   f1  \\
\midrule
Multilingual BERT & 172M & 1.0 & 76.8 & 77.8 & 87.5 & 91.1 & 80.3 & 86.5 & 83.3 \\
XLM-R & 270M & 0.5x & 80.0  & 79.4  & \bf 90.1 & 92.6 & \bf 83.9 & 92.3 & 86.4 \\
KoBERT & 92M & 1.2x & 78.3 & 79.2 & \bf 90.1 &  91.1 & 82.1 & 90.3 & 85.2 \\
ETRI BERT & 110M & - & 79.5 & 80.5 & 88.8 &  93.9 & 82.5 & 94.1 & 86.6 \\
\midrule
\bf KoreALBERT Base & 12M & 5.7x & 79.7 & 81.2 & 89.6 &  93.8 & 82.3 & 92.6 & 86.5 \\
\bf KoreALBERT Large & 18M & 1.3x & \bf 81.1 & \bf 82.1 & 89.7 &  \bf 94.1 & 83.7 & \bf 94.5 & \bf 87.5 \\
\bottomrule
\end{tabularx}
\end{table*}

\subsection{Evaluation}
We fine-tune KoreALBERT for downstream performance evaluations. For comparison, we consider other pretrained BERT-base language models available off-the-shelf.

\subsubsection{Fine-tuning}
In addition to our KoreALBERT, we have downloaded pretrained models available online: multilingual BERT\footnote{https://github.com/google-research/bert}, XLM-R from Facebook AI Research\footnote{https://github.com/facebookresearch/XLM}, KoBERT\footnote{https://github.com/SKTBrain/KoBERT}, and ETRI BERT\footnote{http://aiopen.etri.re.kr/service\_dataset.php}. We optimize respective hyperparameters for each pretrained model before measuring the best and average scores for each model. For all models, we use a batch size of 64 or 128 and from 3 to 5 epochs with a learning rate from $2.0 \times 10^{-5}$ to $5.0 \times 10^{-5}$ and a max-sequence length from 128 to 512. For NER task, we have found out that longer training epochs tend to work better and fine-tuned up to 7 epochs.

\subsubsection{Downstream Tasks}
We consider six downstream NLP tasks detailed below.
\begin{itemize}
\item KorNLI: Korean NLU Dataset includes two downstream tasks. In Korean Natural Language Inference (KorNLI)~\cite{ham2020kornli}, the input is a pair of sentences, a premise and a hypothesis. The fine-tuned model should predict their relationship in one of the three possible labels: entailment, contradiction, and neutral. KorNLI has a total of 950,354 examples. 
\item KorSTS: the second task from Korean NLU is semantic textual similarity (STS) for Korean language. STS requires to predict how semantically similar the two input sentences are on a 0 (dissimilar) to 5 (equivalent) scale. There are 8,628 KorSTS examples in the Korean NLU dataset.
\item Sentiment analysis: we use Naver Sentiment Movie Corpus,\footnote{https://github.com/e9t/nsmc} (NSMC) the biggest Korean movie review dataset, which is collected by the same method that the massive movie review dataset~\cite{maas-etal-2011-learning} proposes. NSMC consists of 200k reviews of shorter than 140 characters that are labeled with human annotations of sentiment.
\item Paraphrase detection (PD): a PD model predicts whether or not a pair of sentences are semantically equivalent. The dataset we consider contains 7,576 examples from a publicly available github repository.\footnote{https://github.com/songys/Question\_pair}
\item Extractive machine reading comprehension (EMRC): EMRC takes in much longer text sequences as an input compared to other tasks. The EMRC model needs to extract the start and end indices inside a paragraph containing the answer of a question. KorQuAD 1.0~\cite{NODE07613668} is a Korean dataset for machine reading comprehension, which is similar to SQuAD 1.0~\cite{squad}. Having exactly the same format as SQuAD, KorQuAD 1.0 comprises 60,407 question-answer pairs.

\item Named entity recognition (NER): NER distinguishes a real-world object such as a person, organization, and place (location) from documents. We use the NER corpus\footnote{http://air.changwon.ac.kr/?page\_id=10} constructed by Naver Corp. and Changwon University in South Korea. The corpus has 14 different types of entities with attached tags \texttt{B/I/-}, denoting multi- or single-word entities as described in Table~\ref{tab:ner-data}.
\end{itemize}

\begin{table}[h]
\begin{center}
\caption{Proportion of the type of entities of NER dataset.}
\label{tab:ner-data}
\begin{tabular}{lcc}
\toprule
\textbf{Category} & \textbf{Tag} & \textbf{Amount} \\
\midrule
NUMBER & NUM & 64,876 \\
CIVILIZATION & CVL & 60,918 \\
PERSON & PER & 48,321 \\
ORGANIZATION & ORG & 45,550 \\ 
DATE & DAT & 33,944 \\
TERM & TRM & 22,070 \\
LOCATION & LOC & 21,095 \\
EVENT & EVT & 17,430 \\ 
ANIMAL & ANM & 6,544 \\
ARTIFACTS\_WORKS & AFW & 6,069 \\ 
TIME & TIM & 4,337 \\
FIELD & FLD & 2,386 \\
PLANT & PLT & 267 \\
MATERIAL & MAT & 252 \\
\bottomrule
\end{tabular}
\end{center}
\end{table}

\subsection{Discussion}
As indicated in Table~\ref{tab:final_perform}, KoreALBERT consistently outperforms M-BERT over all downstream NLU tasks considered. While KoreABLERT has the smallest number of model parameters among all monolingual and multilingual language models compared in this paper, it achieves better results in almost all downstream evaluations. The advantage of having fewer computations of KoreALBERT makes its base model about 5.7 faster than M-BERT and its large model 2.2 faster than XLM-R base at training time.

In NSMC and NER, which are single-sentence classification tasks, KoreALBERT is subpar against XLM-R and KoBERT. For NSMC, KoreALBERT-large cannot produce more discriminnative result than the base model. We suspect the main reason for the performance drop being lack of covering the colloquial usage of words and phrases in our pretraining corpora that mostly consists of more formal style of writings such as news articles and wikipedia. Examples in NSMC seem to use much colloquialism. Also, XLM-R has shown a very good performance on the NER task. Such result is due to the fact that NER does not require much high-level language understanding like multi-sentence discourse coherence.
\section{Conclusion}
We have introduced KoreALBERT, a pre-trained monolingual ALBERT model for Korean language understanding. We have described the details about training KoreALBERT. In particular, we have proposed a word order prediction loss, a new training objective, which is compatible with the original MLM and SOP objectives of ALBERT. KoreALBERT consistently outperforms multi and monolingual baselines on 6 downstream NLP tasks while having much fewer parameters. In our future work, we plan to experiment more comprehensively with the KoreALBERT WOP loss: i) replace token-level switching with word-level switching to improve the difficulty of label prediction; ii) use dynamic token shuffling with varying amount of tokens to be shuffled instead of fixed proportion. We also plan to  investigate how well the proposed WOP loss works with non-agglutinative languages like English.

{
\bibliographystyle{IEEEtran}
\bibliography{paper}
}

\end{document}